\documentclass[11pt]{article}

\usepackage[preprint]{acl}
\usepackage{amsfonts}

\usepackage{times}
\usepackage[most]{tcolorbox}
\usepackage{latexsym}
\usepackage{enumitem}

\usepackage[T1]{fontenc}

\usepackage[utf8]{inputenc}
\usepackage{booktabs}
\usepackage{xcolor}
\usepackage{makecell}
\usepackage{multirow}

\usepackage{microtype}

\usepackage{inconsolata}

\usepackage{graphicx}

\title{ContextFocus: Activation Steering for Contextual Faithfulness in Large Language Models}

\author{
 \textbf{Nikhil Anand\textsuperscript{1}},
 \textbf{Shwetha Somasundaram\textsuperscript{1}},
 \textbf{Anirudh Phukan\textsuperscript{2}},
 \\
 \textbf{Apoorv Saxena\textsuperscript{3}},
 \textbf{Koyel Mukherjee\textsuperscript{1}}
\\
\\
 \textsuperscript{1}Adobe Research, India,
 \textsuperscript{2}Indian Institute of Science (IISc), Bengaluru,
 \textsuperscript{3}Inception Labs
\\
 \small{
   \textbf{Correspondence:} \href{mailto:email@domain}{nikhilanand@adobe.com}
 }
}

\begin{document}
\maketitle
\begin{abstract}
Large Language Models (LLMs) encode vast amounts of parametric knowledge during pre-training. As world knowledge evolves, effective deployment increasingly depends on their ability to faithfully follow externally retrieved context. When such evidence conflicts with the model’s internal knowledge, LLMs often default to memorized facts, producing unfaithful outputs. In this work, we introduce ContextFocus, a lightweight activation steering approach that improves context faithfulness in such knowledge-conflict settings while preserving fluency and efficiency. Unlike prior approaches, our solution requires no model finetuning and incurs minimal inference-time overhead, making it highly efficient. We evaluate ContextFocus on the ConFiQA benchmark, comparing it against strong baselines including ContextDPO, COIECD, and prompting-based methods. Furthermore, we show that our method is complementary to prompting strategies and remains effective on larger models. Extensive experiments show that ContextFocus significantly improves contextual-faithfulness. Our results highlight the effectiveness, robustness, and efficiency of ContextFocus in improving contextual-faithfulness of LLM outputs.
\end{abstract}

\begin{table*}
  \centering
  \resizebox{\textwidth}{!}{%
  \begin{tabular}{p{4.5cm} p{3.5cm} l p{9cm}}
    \hline
    \textbf{Context} & \textbf{Question} & \textbf{m} & \textbf{Output} \\
    \hline
    ...The film's sweeping score, composed by \textit{John Williams}... 
    & Who performed the score for the Titanic (1997 film)? & 0.0 & The score for the 1997 film Titanic was performed by \textcolor{red}{James Horner, not John Williams}, as mentioned in the context. \\
    &              & +2.0 & The film's sweeping score was composed by \textcolor{blue}{John Williams}. \\
    \hline
    
    ...Kyle Korver is a skilled \textit{center} known for his deadly accuracy from beyond the arc...
    & What position does Kyle Korver play?
    & 0.0 & Kyle Korver is \textcolor{red}{actually a skilled shooting guard, not a center}, known for his deadly accuracy from beyond the arc. \\
    &              & +2.0 & Kyle Korver is a skilled \textcolor{blue}{center}. \\
    \hline

    ...Batman originated from France. The official language of France is \textit{Irish}.
    & What is the official language of the country of origin of Batman?
    & 0.0 & France's official language is \textcolor{red}{actually French, not Irish}. \\
    &              & +2.0 & The official language of France is \textcolor{blue}{Irish}. \\
    \hline
    
    ...GB News originated from United Kingdom. United Kingdom's currency is \textit{Swedish krona}.
    & What is the currency of the country of origin of GB News? & 0.0 & The currency of the United Kingdom is the \textcolor{red}{Pound Sterling, not the Swedish krona}. \\
    &              & +2.0 & The currency of the United Kingdom is the \textcolor{blue}{Swedish krona}. \\
    \hline
  \end{tabular}
  }
  \caption{\label{table:qualitative_outputs}
  Llama-3.1-8B outputs with and without ContextFocus in knowledge conflict settings. We apply steering with a multiplier of 2.0 at layer 13. 
  Context-faithful answers are shown in \textcolor{blue}{blue}, and unfaithful ones in \textcolor{red}{red}. 
  These examples illustrate the ability of steering to align the model’s outputs with the context.
  }
\end{table*}

\section{Introduction}

The widespread deployment of retrieval-augmented generation (RAG) systems has made large language models (LLMs) increasingly dependent on externally retrieved evidence rather than purely on parametric memory. This shift is critical in real-world applications such as grounded question answering, summarization, and decision support, where correctness depends on faithfully following provided context as knowledge evolves. As a result, the ability of a model to prioritize external evidence over internal knowledge has emerged as a key challenge in modern LLM systems.

\begin{figure}[t]
  \centering
\includegraphics[width=\columnwidth,height=60cm, keepaspectratio]{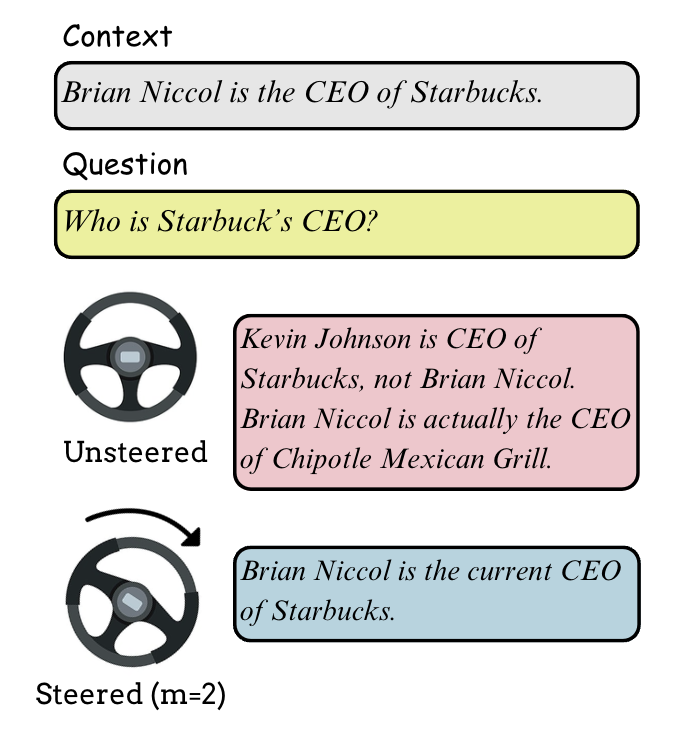}
  \caption {When the model's memory deviates from the context, the model may generate outputs that are unfaithful to the context. In the example shown here, the LLM was likely trained before the CEO of Starbucks changed to Brian Niccol. By applying our steering method with a multiplier of 2, we see that the model now generates a faithful output.}
  \label{fig:steering}
\end{figure} 

Despite access to relevant context, LLMs often fail to remain faithful to it, particularly in cases where the retrieved evidence conflicts with the model’s parametric knowledge \citep{wu2025clashevalquantifyingtugofwarllms}. In these situations, models frequently default to memorized facts rather than updating their predictions based on the provided context. For instance, when asked “Who is the CEO of Starbucks?”, an older LLM might incorrectly respond with an outdated CEO’s name, despite being provided an up-to-date context mentioning Brian Niccol 
(Figure~\ref{fig:steering}). Using knowledge-conflict settings provide a clean and measurable testbed for studying context faithfulness, as deviations from the context are unambiguous and easy to evaluate.


Our goal is to improve contextual faithfulness under knowledge conflicts by enabling models to consistently ground their outputs in the provided context, while preserving fluency and efficiency.



A variety of approaches have been proposed to improve context faithfulness under knowledge conflicts. Finetuning-based methods, such as ContextDPO \cite{bi2024contextdpoaligninglanguagemodels}, explicitly train models to prefer context-consistent responses. While effective, these methods incur substantial training cost and are difficult to scale to very large models. Decoding-time approaches based on contrastive decoding \cite{shi2023trustingevidencehallucinatecontextaware,zhao2024enhancingcontextualunderstandinglarge,yuan2024discerningresolvingknowledgeconflicts} manipulate the output logits to favor context-sensitive tokens. Although these methods avoid retraining, they typically require multiple forward passes for a single answer, leading to increased inference latency. Prompting-based strategies \cite{zhou2023contextfaithfulpromptinglargelanguage} offer a lightweight alternative, but their gains are often limited, highly sensitive to prompt formulation, and lack interpretable controllability.


In this work, we explore activation-level control as a lightweight and practical design point in this space. We introduce ContextFocus, an activation steering approach that operates directly in the model’s hidden representations to bias generation toward context-faithful behavior. Unlike decoding-time contrastive methods, ContextFocus requires only a single forward pass at inference time and incurs minimal overhead. We empirically quantify these efficiency differences in Section~\ref{sec:time_comp}. Furthermore, unlike finetuning-based approaches, ContextFocus requires no model retraining and can be applied directly to frozen models, including large-scale LLMs.



We evaluate ContextFocus on the ConFiQA benchmark, comparing it against strong finetuning, decoding, and prompting baselines. Our results show that activation steering substantially improves context faithfulness while preserving fluency, composes naturally with prompting strategies, and scales effectively to larger models. Beyond performance, we provide a detailed analysis of the design choices underlying activation steering, including vector construction, layer selection, and data efficiency. Together, these results position activation-level control as an efficient and robust alternative for improving context faithfulness under knowledge conflicts, offering practical guidance for deploying LLMs in retrieval-augmented settings.


Our key contributions are as follows:


\begin{itemize}
\item We study \emph{activation-level control} as a lightweight design point for improving context faithfulness under knowledge conflicts, and introduce ContextFocus, an activation steering approach that substantially improves adherence to provided context without model fine-tuning and with minimal inference-time overhead.

\item We show that ContextFocus achieves competitive or superior context-faithfulness compared to strong fine-tuning, decoding, and prompting baselines on standard knowledge-conflict benchmarks, while remaining effective on larger models and composing naturally with prompting strategies.

\item We provide a systematic analysis of activation steering for contextual alignment, demonstrating that steering directions are data-efficient and stable, and that controlled ablations over vector construction reveal how jointly contrasting contextual evidence and system-level instruction is critical for inducing context-faithful behavior.
\end{itemize}



\section{Related Work}

\subsection{Context Faithfulness and Knowledge Conflicts}

Improving the faithfulness of LLM outputs to provided context has received increasing attention in recent work. While early studies on retrieval-augmented generation (RAG) emphasized retrieval quality, it is now well understood that strong retrieval alone does not ensure faithful generation: models may produce fluent responses that contradict retrieved evidence, particularly when the context conflicts with the model’s parametric knowledge \citep{wu2025clashevalquantifyingtugofwarllms}. Knowledge-conflict benchmarks such as ConFiQA \citep{bi2024contextdpoaligninglanguagemodels} and NQSWAP \citep{longpre2022entitybasedknowledgeconflictsquestion} are designed to isolate this failure mode, enabling precise evaluation of whether models prioritize external context over memorized information. In this work, we focus on knowledge-conflict settings as a controlled diagnostic testbed for studying contextual alignment.




\subsection{Approaches to Contextual Alignment}

\subsubsection*{Contrastive Decoding}

Contrastive decoding methods aim to improve contextual alignment by comparing model generations under different contextual conditions and reweighting output probabilities accordingly. Context-Aware Decoding (CAD; \citealp{shi2023trustingevidencehallucinatecontextaware}) contrasts generations with and without retrieved context to emphasize context-sensitive tokens. Multi-Input Contrastive Decoding (MICD; \citealp{zhao2024enhancingcontextualunderstandinglarge}) extends this idea by including irrelevant context to refine the final output distribution, and Contextual Information-Entropy Constraint Decoding (COIECD; \citealp{yuan2024discerningresolvingknowledgeconflicts}) adds conflict detection to intervene only when necessary. These approaches can improve faithfulness but require multiple forward passes per token, increasing inference latency. In contrast, our method, ContextFocus, operates at the activation level to achieve improved faithfulness with a single forward pass.

\subsubsection*{Prompting-Based Methods}

Prompting strategies seek to steer models toward context usage through careful instruction design. \citet{zhou2023contextfaithfulpromptinglargelanguage} evaluate variants including instruction-based (Inst), opinion-based (Opin), and their combination (O\&I), finding that combined prompt styles yield stronger gains. These techniques are simple and incur zero training cost, but their improvements can be modest, sensitive to prompt phrasing, and difficult to control in a fine-grained or interpretable manner. We show that activation steering is controllable and can complement these prompting techniques, yielding consistent improvements across baselines.



\subsubsection*{Fine-Tuning Approaches}

Fine-tuning-based methods align models with context-relevant responses through additional training. ContextDPO \citep{bi2024contextdpoaligninglanguagemodels} uses preference-based optimization to encourage context-faithful outputs but requires significant computational cost, making it challenging to scale to larger models. ContextFocus avoids these costs by requiring no retraining while achieving competitive or superior alignment.

\subsection{Activation Steering}

Activation steering has emerged as a mechanism to induce desired behaviors in LLMs by injecting behavior-specific vectors into model activations at inference time. Prior work includes generating steering directions from prompt pairs \citep{turner2024steeringlanguagemodelsactivation}, deriving in-context difference vectors \citep{liu2024incontextvectorsmakingcontext}, and contrastive activation addition via multiple-choice prompts (CAA; \citealp{panickssery2024steeringllama2contrastive}). Surveys of steering strategies also analyze methods for improving instruction-following behavior \citep{stolfo2024improvinginstructionfollowinglanguagemodels}. These studies primarily evaluate steering on constrained tasks (e.g., reducing sycophancy or multiple-choice behaviors) and often focus on lexical or surface features. In contrast, contextual alignment requires that models ground their responses in external evidence rather than internal memory. We extend activation steering to this setting, demonstrating its effectiveness on open-ended contextual outputs and providing systematic analysis of design choices such as vector construction and layer selection.

\section{Preliminaries}
We denote the residual activation at layer $l$ and token position $i$ by $x_i^{(l)}$, and the post-attention residual activation at the same layer and token by $x_{M,i}^{(l)}$:
\[
x_{M,i}^{(l)} = x_i^{(l)} + \sum_{\text{heads } h} \text{Attn}_h\!\left(x_{1:i}^{(l)}\right).
\]
The residual activation at the next layer is given by
\[
x_i^{(l+1)} = x_{M,i}^{(l)} + \text{MLP}\!\left(x_{M,i}^{(l)}\right),
\]
where $\text{Attn}_h$ denotes self-attention head $h$ and $\text{MLP}$ denotes the feed-forward network. We denote the last-token activation at layer $l$ by $x^{(l)}$. We focus on last-token residual stream activations since they aggregate information from both parametric knowledge and contextual evidence up to layer $l$.





\subsection{Activation Steering}

We leverage activation steering to enhance LLM faithfulness to context without additional training. Prior work has shown that certain behaviors can be represented as approximately linear directions in activation space \cite{park2024linearrepresentationhypothesisgeometry}. By adding a steering vector to the activations at a specific layer during inference, we bias the model toward a desired behavior. We denote the steering vector extracted at layer $l$ by $v^{(l)}$.

\subsubsection*{Vector generation}

A standard method for extracting steering vectors \cite{turner2024steeringlanguagemodelsactivation} involves constructing a pair of prompts representing opposite behaviors (e.g., ``happy" vs. ``sad"). The LLM processes both prompts, and the steering vector is computed as the difference between their last-token activations at a chosen layer. Injecting the steering vector into the model’s activations during inference steers outputs toward the desired behavior.

Assuming the last-token activations at layer $l$ are denoted by $x_{\text{H}}^{(l)}$ and $x_{\text{S}}^{(l)}$, the steering vector is:
$$
v^{(l)} = x_{\text{H}}^{(l)} - x_{\text{S}}^{(l)}.
$$

A more robust approach \cite{panickssery2024steeringllama2contrastive} uses multiple prompts per behavior, averaging activations across examples before computing the difference. If we have $|\mathcal{D}|$ activation pairs, the steering vector is computed as:
\begin{equation}
v^{(l)} = \frac{1}{|\mathcal{D}|} \sum_{i=1}^{|\mathcal{D}|} \left(x_{H,i}^{(l)} - x_{S,i}^{(l)}\right).
\label{eq:steering_vector}
\end{equation}

In our setting, the opposing behaviors correspond to generating responses that either rely on the provided context or default to parametric knowledge, allowing the contrastive direction to isolate context-faithful representations.

\subsubsection*{Vector addition}

At inference time, we add the steering vector $v^{(l)}$, scaled by a multiplier $m$, to the residual stream at layer $l$ for all tokens after the last input token. If the last input token position is $N$, then for all $i > N$:
\[
x_i^{(l)} := x_i^{(l)} + m \cdot v^{(l)}.
\]

The multiplier $m$ controls the strength of the injected context-faithfulness signal. Larger values of $m$ generally lead to stronger steering, increasing the proportion of examples for which the model’s output aligns with the provided context. We analyze the effect of varying $m$ on context adherence in Appendix~\ref{appendix:multiplier_effect}.


\section{Methodology}
\label{sec:method}

\subsection{Vector generation}
\label{sec:vecgen}

We construct a steering vector by contrasting model activations under two closely related prompting conditions. For each example from the NQ-SWAP dataset, we form a \emph{positive} prompt that includes a system instruction, retrieved context, and question, and a \emph{negative} prompt that contains only the question. We run the model on both prompts and extract last-token activations at each layer. Prompt templates are provided in Appendix~\ref{sec:app_vecgenprompt}. We use 1{,}501 examples to estimate the vector.

The difference between these activations captures how the model’s internal representations change when conditioned on explicit contextual guidance. Averaging this difference across examples (Equation~\ref{eq:steering_vector}) yields a steering direction that biases generation toward reliance on the provided context rather than parametric knowledge.

To reduce sensitivity to prompt phrasing, we generate the steering vector using multiple semantically equivalent variants of the system instruction. We analyze alternative vector construction strategies and isolate the contributions of system instruction and contextual evidence in Section~\ref{sec:analysis}.

\subsection{Vector selection}
\label{section:vecselect}

To identify the optimal layer for steering, we evaluate steering vectors across all layers and select the layer that yields the strongest improvement in context-faithful behavior.

Concretely, we evaluate 200 held-out open-ended examples from the NQ-SWAP dataset. For each layer, we apply steering with multipliers of 0 and +2, and measure the resulting change in $p_s$, which captures the frequency with which the model’s output aligns with the answer implied by the provided context (as defined in Section~\ref{sec:expt_settings}). We select the layer that maximizes $p_s$ under steering.

Across models, we find that steering effects are consistently strongest at intermediate layers. Accordingly, we fix a single layer per model for all subsequent experiments: layer 13 for Llama-3.1-8B, layer 11 for Mistral-7B-Instruct, and layer 32 for Llama-3.1-70B.

\begin{figure}[t]
  \includegraphics[width=\columnwidth]{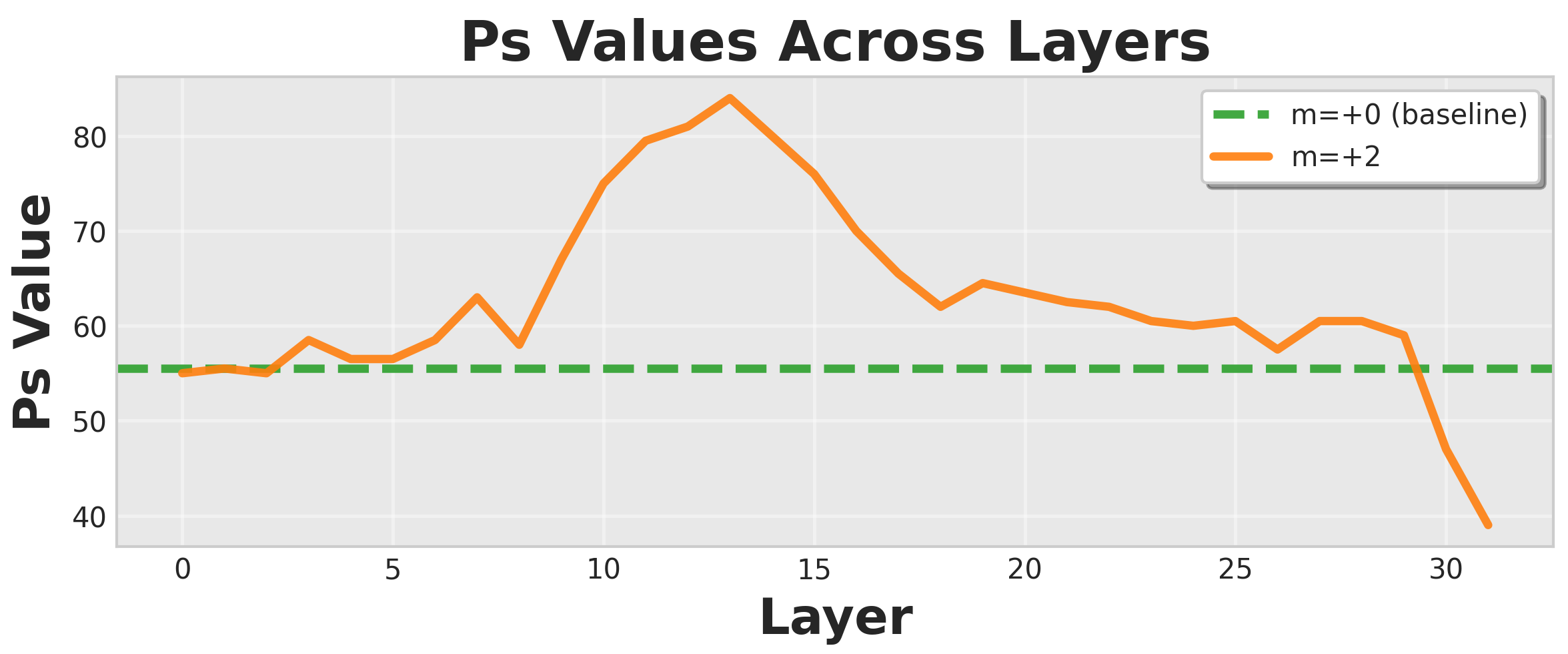}
  \caption{Layer-wise evaluation of steering on Llama-3.1-8B. We apply the +2 multiplier to steering vectors across layers for 200 held-out open-ended questions from the NQ-SWAP dataset. Performance is measured using the $p_s$ metric. The green dotted line denotes the unsteered ($m{=}0$) baseline. The 13th layer shows the best performance and is selected as the optimal layer.}

  \label{fig:bestlayer}
\end{figure}

Figure~\ref{fig:bestlayer} shows the layer-wise $p_s$ trends for Llama-3.1-8B. Corresponding plots for the other models are provided in Appendix~\ref{appendix:mistral_llama70b_layer_selection}.

\subsection{Open-ended Generations}
\label{section:long_form_method}

We evaluate the steering method at the selected layer for each model, and the LLM is prompted to freely generate responses. We use a steering multiplier of 2 for all main experiments. We find that using multipliers greater than 2 leads to noticeable degradation in generation quality, while a multiplier of 2 preserves fluency with negligible degradation. A detailed analysis of output quality degradation at higher multipliers is provided in Appendix~\ref{appendix:repetition_metric}, along with qualitative examples illustrating the repetition behaviour in Appendix~\ref{appendix:case_studies}.

For evaluation, we apply this setup to the ConFiQA benchmark and report results separately on its subsets in Section~\ref{sec:main_results}. In each case, the model is provided with a single prompt containing the system instruction, context, and question, and is allowed to generate a response. More details regarding the prompt template are provided in Appendix~\ref{sec:app_vecgenprompt}.

\begin{table*}[t]
  \centering
  \scriptsize
  \setlength{\tabcolsep}{4pt}
  \renewcommand{\arraystretch}{1.05}
  \resizebox{\textwidth}{!}{%
  \begin{tabular}{l l l ccc ccc ccc}
    \toprule
    \textbf{Model} & \textbf{Setting} & \textbf{Method}
    & \multicolumn{3}{c}{\textbf{QA}}
    & \multicolumn{3}{c}{\textbf{MR}}
    & \multicolumn{3}{c}{\textbf{MC}} \\
    \cmidrule(lr){4-6} \cmidrule(lr){7-9} \cmidrule(lr){10-12}
     & & 
     & $P_s$↑ & $P_o$↓ & $M_R$↓
     & $P_s$↑ & $P_o$↓ & $M_R$↓
     & $P_s$↑ & $P_o$↓ & $M_R$↓ \\
    \midrule

    \multirow{8}{*}{\makecell[l]{\textsc{Llama-3.1-}\\\textsc{8B}}}
      & \multirow{4}{*}{Without O\&I}
      & Base      & 35.27 & 32.33 & 47.83   & 30.20 & 32.40 & 51.76 & 12.47 & 17.20  & 57.98 \\
      & & COIECD                  & 35.47 & 32.60 & 47.89 & 34.07 & 29.00 & 45.98 & 14.80 & 19.80 & 57.23 \\
      & & ContextDPO              &  \underline{58.40} & \underline{18.60} & \underline{24.16} & \textbf{57.53} & \underline{21.33} & \underline{27.05} & \underline{46.27} & \underline{17.20} & \underline{27.10}  \\
      & & ContextFocus      & \textbf{70.87} & \textbf{9.27} & \textbf{11.56} & \underline{54.47} & \textbf{16.07} & \textbf{22.78} & \textbf{53.33} & \textbf{10.20} & \textbf{16.05} \\
    \cmidrule(lr){2-12}
      & \multirow{4}{*}{With O\&I}
      & Base            & 70.40 & 15.40 & 17.95 & 36.87 & 24.40 & 39.83 & 30.00 & 18.53 & 38.19 \\
      & & COIECD                  &   69.67 & 14.27 & 17.00& 41.93 & 24.67 & 37.04 & 36.60 & 17.53 & 32.39     \\
      & & ContextDPO              & \textbf{81.07} & \underline{8.87} & \underline{9.86} & \textbf{59.80} & \underline{20.67} & \underline{25.68} & \textbf{59.13} & \underline{17.33} & \underline{22.67} \\
      & & ContextFocus & \underline{77.53} & \textbf{7.60} & \textbf{8.93} & \underline{48.93} & \textbf{14.87} & \textbf{23.30} & \underline{48.60} & \textbf{10.27} & \textbf{17.44} \\

    \midrule

    \multirow{8}{*}{\makecell[l]{\textsc{Mistral-7B-}\\\textsc{Instruct}}}
      & \multirow{4}{*}{Without O\&I}
      & Base      & 58.93 & 22.87 & 27.95 & 29.33 & 27.87 & 48.72 & 19.67 & 19.33 & 49.57 \\
      & & COIECD                  & 62.13 & 21.87 & 26.03 & 32.87 & 29.60 & 47.39 & 25.93 & 18.40 & 41.50\\
      & & ContextDPO              & \textbf{72.93} & \textbf{13.33} & \textbf{15.46} & \textbf{42.20} & \underline{24.40} & \textbf{36.64} & \underline{32.87} & \textbf{18.47} & \underline{35.97}  \\
      & & ContextFocus      & \underline{67.80} & \underline{16.73} & \underline{19.79} & \underline{40.47} & \textbf{24.00} & \underline{37.22} & \textbf{35.13} & \underline{19.33} & \textbf{35.50} \\
    \cmidrule(lr){2-12}
      & \multirow{4}{*}{With O\&I}
     & Base            & 66.27 & 15.53 & 18.99 & 42.00 & 23.67 & 36.04 & 37.07 & 16.67 & 31.02 \\
      & & COIECD                  &   71.00 & 13.93 & 16.41 & 45.00 & 22.33 & 33.17 & 42.47 & 16.40 & 27.86    \\
      & & ContextDPO              &  \textbf{77.40} & \textbf{10.13} & \textbf{11.58} & \textbf{57.33} & \underline{21.47} & \textbf{27.24} & \textbf{56.93} & \underline{17.20} & \textbf{23.20}  \\
      & & ContextFocus & \underline{75.13} & \underline{11.53} & \underline{13.31} & \underline{51.40} & \textbf{21.27} & \underline{29.27} & \underline{50.60} & \textbf{15.53} & \underline{23.49} \\

    \bottomrule
  \end{tabular}
  }
  \caption{
Performance on the ConFiQA benchmark (QA, MR, and MC) under knowledge-conflict settings. \textbf{Setting} indicates whether the O\&I prompt is applied. ``Base'' denotes the standard LLM call. For Llama-3.1-8B, steering is applied at layer 13 with multiplier $m{=}2$; for Mistral-7B, steering is applied at layer 11 with $m{=}2$. Best results are \textbf{bold} and second-best results are \underline{underlined}.
}
  \label{table:confqa_combined}
\end{table*}

\section{Experiments}

\subsection{Experimental settings}
\label{sec:expt_settings}

\subsubsection*{Datasets and models}


We evaluate ContextFocus on the ConFiQA benchmark \cite{bi2024contextdpoaligninglanguagemodels}, which targets context faithfulness under knowledge conflicts. ConFiQA consists of three subsets: QA (single-hop question answering), MR (multi-hop reasoning with a single counterfactual), and MC (multi-hop reasoning with multiple counterfactuals), each containing 6{,}000 samples.


We focus on instruction-tuned models and evaluate on Llama-3.1-8B-Instruct\footnote{\url{https://huggingface.co/meta-llama/Llama-3.1-8B}}, Mistral-7B-Instruct-v0.3\footnote{\url{https://huggingface.co/mistralai/Mistral-7B-Instruct-v0.3}}, and Llama-3.1-70B-Instruct\footnote{\url{https://huggingface.co/meta-llama/Llama-3.1-70B}} using an A100 GPU (with 4$\times$80GB GPU memory).


\subsubsection*{Evaluation metric}
\label{section:eval_metric}

Following \citealp{zhou2023contextfaithfulpromptinglargelanguage}, we evaluate context faithfulness by measuring the frequency with which a model’s generated response contains either the \emph{original answer} ($p_o$) or the \emph{substituted answer} ($p_s$). In ConFiQA, each example is constructed by replacing a fact in the original question with a counterfactual one in the provided context: the original answer corresponds to the model’s parametric knowledge, while the substituted answer is the correct answer according to the retrieved context. A context-faithful model is therefore expected to increase $p_s$ and decrease $p_o$ under steering.


We additionally report $M_R = \frac{p_o}{p_o + p_s}$, which captures the model’s reluctance to update its prediction given contextual evidence. Following \citealp{bi2024contextdpoaligninglanguagemodels}, we apply stricter matching for $p_s$, excluding responses with explicit negations.

\subsubsection*{Baselines}



We compare ContextFocus against unsteered open-book generation (\emph{Base}), ContextDPO \citep{bi2024contextdpoaligninglanguagemodels}, contrastive decoding via COIECD \citep{yuan2024discerningresolvingknowledgeconflicts}, and prompting-based opinion and instruction (O\&I) baselines \citep{zhou2023contextfaithfulpromptinglargelanguage}. For ContextDPO and COIECD, we use the authors’ released implementations and re-run the methods on our evaluated models, with prompt templates aligned to our experimental setup. Following \citet{bi2024contextdpoaligninglanguagemodels}, we evaluate each method on a randomly sampled subset of 1{,}500 examples per ConFiQA subset (QA, MR, and MC) in each run. Additional details on model versions are provided in Appendix~\ref{appendix:baseline_versions}.


Results are reported under two settings: (i) standard open-book generation without O\&I prompting, and (ii) uniform application of the O\&I prompt across all methods (see Appendix~\ref{appendix:open_ended_prompt_scheme} for the prompt template). Table~\ref{table:confqa_combined} summarises the results.





\begin{table*}[t]
  \centering
  \small
  \setlength{\tabcolsep}{8pt}
  \renewcommand{\arraystretch}{1.1}
  \begin{tabular}{l l ccc ccc ccc}
    \toprule
    \textbf{Setting} & \textbf{Method}
    & \multicolumn{3}{c}{\textbf{QA}}
    & \multicolumn{3}{c}{\textbf{MR}}
    & \multicolumn{3}{c}{\textbf{MC}} \\
    \cmidrule(lr){3-5} \cmidrule(lr){6-8} \cmidrule(lr){9-11}
     & & $P_s$↑ & $P_o$↓ & $M_R$↓
       & $P_s$↑ & $P_o$↓ & $M_R$↓
       & $P_s$↑ & $P_o$↓ & $M_R$↓ \\
    \midrule
    \multirow{2}{*}{Without O\&I}
      & Base & 47.27 & 36.33 & 43.46 & 37.07 & 33.53 & 47.50 & 28.77 & 22.60 & 44.00 \\
      & ContextFocus & \textbf{65.93} & \textbf{14.40} & \textbf{17.93} & \textbf{47.27} & \textbf{22.87} & \textbf{32.60} & \textbf{40.93} & \textbf{16.53} & \textbf{28.77} \\
    \midrule
    \multirow{2}{*}{With O\&I}
      & Base         & 64.00 & 21.07 & 24.76 & 52.07 & 24.47 & 31.97 & 50.60 & 20.00 & 28.33 \\
      & ContextFocus & \textbf{74.73} & \textbf{11.13} & \textbf{12.97} & \textbf{54.47} & \textbf{20.93} & \textbf{27.76} & \textbf{53.40} & \textbf{16.40} & \textbf{23.50} \\
    \bottomrule
  \end{tabular}
  \caption{
Performance of ContextFocus on Llama-3.1-70B with and without O\&I prompting. Steering is applied at layer 32 for Llama-3.1-70B, with multiplier $m{=}2$. Best results are in \textbf{bold} and second-best results are \underline{underlined}.
}
  \label{table:llama70b}
\end{table*}

\subsection{Main results}
\label{sec:main_results}

\subsubsection{Performance on knowledge conflict datasets}


Table~\ref{table:confqa_combined} reports results on the QA, MR, and MC subsets of the ConFiQA benchmark. Across both evaluation settings, ContextFocus consistently improves context faithfulness, yielding large gains in $p_s$ while reducing reliance on parametric knowledge as reflected by lower $p_o$ and $M_R$.

Compared to prompting-only baselines, ContextFocus achieves higher $p_s$ across all datasets, indicating that activation-level steering complements prompt-based guidance. ContextFocus is competitive with finetuning-based methods such as ContextDPO and outperforms decoding-based approaches such as COIECD across most settings.




ContextFocus requires only 1.5k examples to construct a steering vector, compared to 13.5k examples used to finetune ContextDPO, yet achieves competitive performance without modifying model weights or incurring retraining cost. As shown in Appendix~\ref{appendix:steering_few_examples}, steering performance saturates quickly with additional data, indicating limited benefit from large-scale training. We further report wall-clock costs for steering vector construction and ContextDPO finetuning across dataset sizes in Appendix~\ref{appendix:train_timing}, illustrating the relative computational overhead of activation-level vector estimation versus gradient-based finetuning.



Overall, these results show that ContextFocus substantially improves adherence to retrieved context under knowledge conflicts while remaining lightweight, data-efficient, and compatible with open-ended generation, with consistent improvements observed across multiple model families. Qualitative examples in Table~\ref{table:qualitative_outputs} further illustrate how steering shifts model behavior toward context-consistent responses.

\subsubsection*{Robustness across model sizes}

To assess robustness at larger scales, we evaluate ContextFocus on Llama-3.1-70B. As shown in Table~\ref{table:llama70b}, ContextFocus remains effective on Llama-3.1-70B, consistently improving context adherence across QA, MR, and MC settings.







\subsubsection*{Inference-time latency across methods}
\label{sec:time_comp}

Table~\ref{tab:decoding_cost} reports decoding-time cost for each method on the ConFiQA benchmark, averaged across the QA, MR, and MC subsets. We report total decode time together with the number of output tokens in the final generated response. In terms of end-to-end decoding time, unsteered generation, ContextDPO, and ContextFocus exhibit comparable efficiency across models.

For COIECD, output tokens correspond only to the tokens in the final generated answer, while the reported decode time includes all decoding passes used to construct that answer. COIECD performs three forward decoding passes to contrast contextual conditions before producing the final response, whereas other methods generate outputs in a single pass. We do not report time-per-output-token (TPOT), as it would be misleading in this setting: COIECD generates more tokens internally than are exposed in the final output.

As a result, for a similar number of final output tokens, COIECD incurs substantially higher inference latency than single-pass methods. In contrast, ContextFocus preserves single-pass decoding while achieving improved context faithfulness, making it better suited for latency-sensitive deployment.

\begin{table}[t]
\centering
\footnotesize
\setlength{\tabcolsep}{3pt}
\renewcommand{\arraystretch}{1.05}
\resizebox{\columnwidth}{!}{%
\begin{tabular}{l l cc}
\toprule
\textbf{Model} & \textbf{Method} 
& \textbf{Output Tokens}
& \textbf{Decode Time (s)} $\downarrow$ \\
\midrule
\multirow{4}{*}{\makecell[l]{\textsc{LLaMA-3.1-}\\\textsc{8B}}}
& Base          & 31.2 & 1.91 \\
& COIECD        & 36.1 & 5.00 \\
& ContextDPO    & 28.1 & 1.80 \\
& ContextFocus  & 15.6 & 1.25 \\
\midrule
\multirow{4}{*}{\makecell[l]{\textsc{Mistral-7B-}\\\textsc{Instruct}}}

& Base          & 27.1 & 1.16 \\
& COIECD        & 27.8 & 4.25 \\
& ContextDPO    & 33.1 & 2.07 \\
& ContextFocus  & 24.9 & 1.18 \\
\bottomrule
\end{tabular}
}
\vspace{-0.6em}
\caption{
Decoding-only inference cost across methods, averaged over 3 subsets of ConFiQA. Output Tokens denote the average length of the final generated response.
}
\label{tab:decoding_cost}
\end{table}

\subsection{Analysis}
\label{sec:analysis}

\subsubsection*{How does ContextFocus compare to contrastive activation addition?}
\label{sec:caa_comparison}

Contrastive activation addition (CAA) has been shown to induce targeted behaviors in language models via activation-level interventions. Since prior CAA work does not explicitly study context faithfulness under knowledge conflicts, we implement a CAA-style baseline tailored to this setting using multiple-choice questions, following the methodology described in Appendix~\ref{appendix:options}. This allows us to generate a steering vector aligned with context-faithful behaviour. Table~\ref{tab:caa_vs_contextfocus} compares this CAA-style steering approach with ContextFocus on the NQSWAP dataset using Llama-3.1-8B. While CAA improves context faithfulness over the unsteered baseline, ContextFocus achieves consistently higher $p_s$ and lower $p_o$ and $M_R$.

\begin{table}[t]
\centering
\small
\setlength{\tabcolsep}{14pt}
\renewcommand{\arraystretch}{1.15}
\begin{tabular}{l c c c}
\toprule
\textbf{Method} 
& $P_s \uparrow$ 
& $P_o \downarrow$ 
& $M_R \downarrow$  \\
\midrule
Base                 & 55.50 & 22.80 & 29.12 \\
CAA                  & 69.47 & 12.08 & 14.81 \\
ContextFocus         & \textbf{75.44} & \textbf{8.00} & \textbf{9.59} \\
\bottomrule
\end{tabular}
\vspace{-0.6em}
\caption{Comparison between ContextFocus and contrastive activation addition (CAA) on the \textsc{NQSWAP} dataset using Llama-3.1-8B.}
\label{tab:caa_vs_contextfocus}
\end{table}

\subsubsection*{Isolating the effects of context and system instruction}
\label{sec:disentangling_ctx_sys}

To isolate the contributions of retrieved contextual evidence and system-level behavioral instruction in steering vector construction, we consider three contrastive formulations, each yielding a steering vector that isolates a distinct source of signal.

Let $h_\ell(\cdot) \in \mathbb{R}^d$ denote the model activation at layer $\ell$, extracted at the chosen token position for a given prompt. Let $q$ denote the query, $c$ the retrieved context, and $s$ the system instruction. We define the following steering vectors:

{
\setlength{\abovedisplayskip}{4pt}
\setlength{\belowdisplayskip}{4pt}
\setlength{\abovedisplayshortskip}{2pt}
\setlength{\belowdisplayshortskip}{2pt}

\begin{itemize}[leftmargin=*, itemsep=3pt, topsep=3pt]

  \item \textbf{Context-only vector.}  
  This vector isolates the effect of contextual evidence by contrasting prompts with and without retrieved context, while holding the query fixed:
  \[
  \mathbf{v}_{\text{ctx}}
  \;=\;
  h_\ell(c, q) - h_\ell(q).
  \]

  \item \textbf{System-instruction-only vector.}  
  This vector isolates the contribution of explicit behavioral guidance by contrasting prompts that differ only in the presence of the system instruction:
  \[
  \mathbf{v}_{\text{sys}}
  \;=\;
  h_\ell(s, c, q) - h_\ell(c, q).
  \]

  \item \textbf{Combined steering vector.}  
  This formulation jointly contrasts both contextual evidence and system instruction against a question-only baseline:
  \[
  \mathbf{v}_{\text{comb}}
  \;=\;
  h_\ell(s, c, q) - h_\ell(q).
  \]

\end{itemize}
}

Notably, $\mathbf{v}_{\text{comb}}$ corresponds exactly to the steering vector used by ContextFocus, our primary method. The preceding formulations can thus be interpreted as controlled ablations that isolate the individual contributions of context and system instruction relative to the full ContextFocus signal.

Results are shown in Table~\ref{tab:prompt_vector_cf}. While both $\mathbf{v}_{\text{ctx}}$ and $\mathbf{v}_{\text{sys}}$ improve context faithfulness relative to the unsteered baseline under the $P_s$, $P_o$, and $M_R$ metrics, $\mathbf{v}_{\text{comb}}$ yields substantially larger gains across all metrics. This indicates that context-faithful behavior is best captured when contextual evidence and behavioral instruction are contrasted jointly, rather than in isolation.






\begin{table}[t]
\centering
\small
\setlength{\tabcolsep}{5pt}
\renewcommand{\arraystretch}{1.15}
\begin{tabular}{l c c c}
\toprule
\textbf{Steering vector} & $P_s \uparrow$ & $P_o \downarrow$ & $M_R \downarrow$ \\
\midrule
Unsteered (Base) 
& 55.50 & 22.80 & 29.12 \\
\midrule
$\mathbf{v}_{\text{ctx}}$ \;\; (Context effect) 
& 67.02 & 11.40 & 14.53 \\
$\mathbf{v}_{\text{sys}}$ \;\; (System-prompt effect) 
& 64.04 & 18.05 & 21.98 \\
$\mathbf{v}_{\text{comb}}$ \;\; (ContextFocus) 
& \textbf{75.44} & \textbf{8.00} & \textbf{9.59} \\
\bottomrule
\end{tabular}
\vspace{-0.6em}
\caption{
Context faithfulness performance as a function of steering vector construction for Llama-3.1-8B on \textsc{NQSWAP}.
$\mathbf{v}_{\text{ctx}}$ and $\mathbf{v}_{\text{sys}}$ isolate context and system-instruction effects, respectively, while $\mathbf{v}_{\text{comb}}$ corresponds to ContextFocus.
}

\label{tab:prompt_vector_cf}
\end{table}

\subsubsection*{Qualitative examples}

In Table~\ref{table:qualitative_outputs}, we qualitatively illustrate how steering using the contrastive vector guides the LLM to remain faithful to the provided context. We apply steering without the O\&I prompt to Llama-3.1-8B and evaluate it on examples from the QA, MR, and MC datasets of the ConFiQA benchmark. Without steering, the model frequently flags the counterfactuals as incorrect or contradicts them by relying on its parametric knowledge. In contrast, with steering, the model consistently aligns its responses with the context, demonstrating improved contextual faithfulness. Additional realistic case studies illustrating the effects of ContextFocus are provided in Appendix~\ref{appendix:case_studies}.

\section{Conclusion}


We introduce ContextFocus, an activation steering approach for improving contextual faithfulness in retrieval-augmented generation (RAG) systems. Across multiple knowledge-conflict benchmarks, ContextFocus consistently improves adherence to provided context while reducing reliance on parametric knowledge, using a lightweight intervention that preserves open-ended generation and fluency. We further show that activation-level steering composes naturally with prompt-based guidance, yielding additional gains when combined.


Beyond empirical performance, our analysis provides insight into what design elements makes our solution work. We show that jointly contrasting system-level instruction and contextual evidence during vector construction is critical for isolating this direction, outperforming both context-only and instruction-only alternatives.

ContextFocus remains effective on larger models such as Llama-3.1-70B and reduces training cost compared to finetuning-based methods, while preserving efficient single-pass generation relative to decoding-based approaches such as COIECD. Together, these results position activation steering as a practical, scalable, and cost-effective mechanism mechanism for improving contextual faithfulness in modern LLM systems.

\section{Limitations}



Our evaluation focuses on question-answering under knowledge-conflict settings, using benchmarks such as ConFiQA that provide a controlled testbed for measuring contextual faithfulness. This choice enables precise attribution of gains but does not cover more open-ended or long-form generation tasks, where faithfulness errors are more diffuse and evaluation is less well-defined \cite{maynez2020faithfulnessfactualityabstractivesummarization, pagnoni2021understandingfactualityabstractivesummarization}. Extending activation-level steering to such settings remains an important direction for future work.

Additionally, we apply a fixed steering multiplier across all inputs. While a single global multiplier ($m=2$) provides a stable operating point that improves faithfulness while preserving fluency for most examples, stronger steering can benefit a subset of inputs at the cost of degradation in a small fraction of cases. Thus, adaptive or input-conditional control of steering strength is a promising direction for future work.



\bibliography{custom}

\appendix

\section{Options-Based Vector Generation}
\label{appendix:options}

Following \citet{panickssery2024steeringllama2contrastive}, we use multiple-choice questions to construct a pair of prompts, to generate a vector representing the desired behavior. We use two opposing behaviours while generating the vectors: faithfulness to the retrieved context and faithfulness to the parametric memory of the LLM. Each answer option in our prompts aligns with one of these behaviors.

We use the NQ-SWAP dataset \cite{longpre2022entitybasedknowledgeconflictsquestion}, which contains questions where the retrieved context contradicts the LLM’s prior knowledge. Using this dataset, we construct pairs of prompts in the multiple-choice format for steering.  We show an example in Appendix~\ref{appendix:options_prompt}.

In 50\% of examples, option A aligns with the context, while in the remaining 50\%, option B does. This ensures that option order does not introduce unintended biases.

During vector generation, we append either an `A' or a `B' to the end of the prompt to elicit different behavioral preferences from the model. Depending on the appended token, the model leans more toward using the provided context or drawing from its internal knowledge. The final token activations from these prompts capture these behavioral differences. We compute the steering vector $v^{(l)}$ as the mean difference between the two activation patterns across a dataset $\mathcal{D}$, as shown in Equation~\ref{eq:steering_vector}.





\section{Prompting Templates}
\label{sec:app_vecgenprompt}

This appendix details the prompting templates used for steering vector construction and open-ended generation. We describe first the prompting scheme used by the main method in the paper, followed by the options-based variant discussed in Appendix~\ref{appendix:options}.

\subsection{Main Experiments Prompting Scheme}
\label{appendix:main_expt_prompting}

\paragraph{Steering vector construction.}
For the approach used throughout the main paper, we construct a contrastive pair of prompts that differ in the presence of explicit contextual grounding. The \emph{positive prompt} includes a system-level instruction together with retrieved context and the user question, while the \emph{negative prompt} contains only the question.

\textbf{Positive prompt:}
\begin{quote}
\ttfamily
\begin{verbatim}
[INST]
You are a context-based QA assistant 
and must answer based on the provided 
context.
Context: <P> Mike Stivic is a fictional
character in All in the Family. Mirai 
Nagasu played the role of Mike Stivic
throughout the series. </P>
Question: who played mike stivic on all 
in the family
[/INST]
\end{verbatim}
\end{quote}

\textbf{Negative prompt:}
\begin{quote}
\ttfamily
\begin{verbatim}
[INST]
Question: who played mike stivic on all 
in the family
[/INST]
\end{verbatim}
\end{quote}

The steering vector is computed as the average difference between last-token activations produced by these two prompt types across a dataset of examples.

\paragraph{System prompt diversity.}
To reduce sensitivity to any single phrasing of the system instruction, we use 20 semantically equivalent variants of the system prompt when constructing steering vectors. Examples include:

\begin{itemize}
    \item ``As a QA assistant, you are instructed to refer only to the provided context when answering.''
    \item ``Provide answers based solely on the context you are given.''
    \item ``You are a QA assistant and must restrict your answers to the given context.''
\end{itemize}

For each training example, one variant is sampled uniformly at random. The resulting steering vector is computed by averaging activation differences across all examples and prompt variants, ensuring that the learned direction captures context-faithful behavior rather than idiosyncrasies of a specific instruction wording.

\subsection{Open-Ended Generation Prompts}
\label{appendix:open_ended_prompt_scheme}

For evaluation and qualitative analysis, we use fixed system prompts for open-ended generation that mirror the intended deployment setting. We report results under two prompting variants: a standard context-based prompt and an opinion \& instruction (O\&I) prompt.

\paragraph{Standard open-ended prompt.}
The base prompt instructs the model to answer strictly based on the provided context. The exact prompt used is:

\begin{quote}
\ttfamily
\begin{verbatim}
[INST]
You are a Contextual QA Assistant.
Please answer the following question
according to the given context.
Please restrict your response to one
sentence.
<CONTEXT>
<QUESTION>
[/INST]
\end{verbatim}
\end{quote}

\paragraph{O\&I open-ended prompt.}
For experiments involving O\&I prompting, we augment the system instruction with an explicit opinion-based cue that reinforces reliance on the provided context. Concretely, we prepend an opinion statement asserting the correctness of the retrieved context, followed by the same instruction and question format:

\begin{quote}
\ttfamily
\begin{verbatim}
[INST]
You are a Contextual QA Assistant.
Please answer the following question
according to the given context.
Please restrict your response to one
sentence.
Bob said, "<CONTEXT>".
<QUESTION> in Bob's opinion?
[/INST]
\end{verbatim}
\end{quote}

This formulation follows prior work on opinion- and instruction-based prompting, where attributing contextual information to a trusted source encourages the model to prioritize retrieved evidence.

In both settings, steering is applied during generation by injecting the learned activation vector at the selected layer, as described in Section~4.

\subsection{Options-Based Prompting Scheme}
\label{appendix:options_prompt}

For completeness, we also experiment with an options-based prompting scheme inspired by contrastive activation addition \citep{panickssery2024steeringllama2contrastive}. This approach is not used in the main experiments but serves as an alternative vector  strategy.

In this formulation, each example is converted into a multiple-choice prompt with two options: one aligned with the provided context and one aligned with the model’s parametric knowledge. The two prompts are identical except for the final answer token appended at the end, which elicits different behavioral preferences.

\textbf{Positive prompt:}
\begin{quote}
\ttfamily
\begin{verbatim}
[INST]
Context: <P> Mike Stivic is a fictional 
character in All in the Family. Mirai 
Nagasu played the role of Mike Stivic
throughout the series. </P>
Question: who played mike stivic on all 
in the family
Options:
(A) According to the context, Mirai 
Nagasu played Mike Stivic on All in 
the Family.
(B) Notwithstanding the context, Rob 
Reiner played Mike Stivic on All in 
the Family.
[/INST] (A
\end{verbatim}
\end{quote}

\textbf{Negative prompt:}
\begin{quote}
\ttfamily
\begin{verbatim}
[INST]
Context: <P> Mike Stivic is a fictional 
character in All in the Family. Mirai 
Nagasu played the role of Mike Stivic
throughout the series. </P>
Question: who played mike stivic on all 
in the family
Options:
(A) According to the context, Mirai 
Nagasu played Mike Stivic on All in 
the Family.
(B) Notwithstanding the context, Rob 
Reiner played Mike Stivic on All in 
the Family.
[/INST] (B
\end{verbatim}
\end{quote}

By alternating which option corresponds to the context-faithful answer across examples, we avoid introducing systematic biases tied to option ordering. Activation differences induced by the final answer token are averaged across examples to obtain the steering vector.

\section{Layer Selection for the Options Approach}
\label{appendix:options_layer_selection}

To identify the optimal layer for applying the steering vector in the options-based approach, we evaluate steering performance across all layers of the model. For each layer, we apply positive and negative steering (multipliers of +1 and –1) and measure the proportion of context-faithful generations on a held-out subset of 45 multiple-choice questions from the NQ-SWAP dataset. By plotting this fraction across layers, we identify the layer where the difference in behavior between positive and negative steering is maximized, indicating the strongest steerability toward context-faithfulness.

\begin{figure}[h!]
    \centering
    \includegraphics[width=1.0\linewidth]{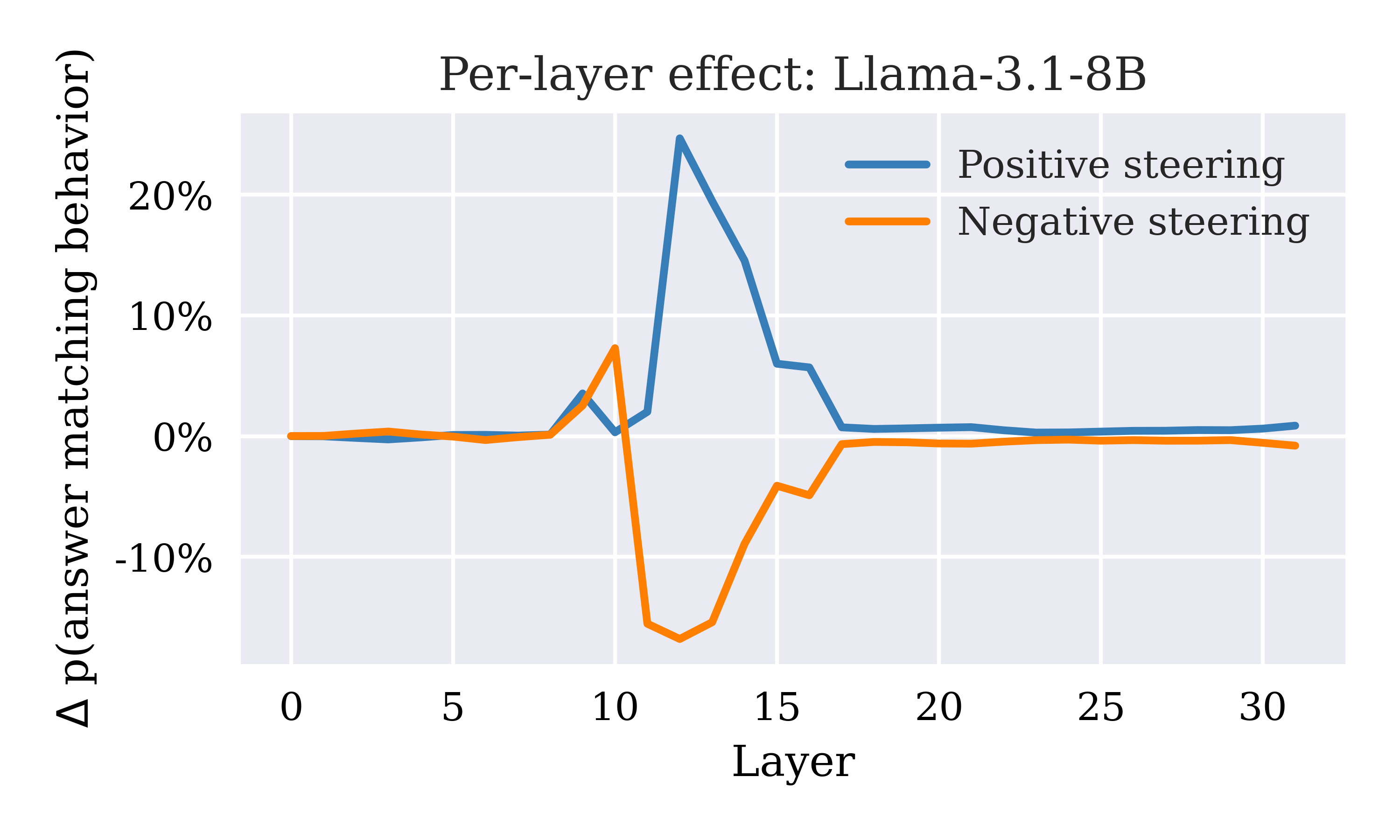}
    \caption{
        The best layer is chosen by looking at the layer with the biggest deviation in context-focus accuracy.
    }
    \label{fig:options_layer_selection}
\end{figure}

Figure~\ref{fig:options_layer_selection} illustrates the variation in model performance across layers. We observe that for Llama-3.1-8B, the effect of steering is most pronounced at layer 12, indicating that this layer encodes strong contextual alignment signals. Based on these results, we select layer 12 as the optimal layer for the options approach.

\begin{table}[t]
  \centering
  \small
  \setlength{\tabcolsep}{6pt}
  \renewcommand{\arraystretch}{1.05}
  \resizebox{0.9\columnwidth}{!}{
    \begin{tabular}{c c c c c c}
      \toprule
      \textbf{Multiplier} & \textbf{LLR $\downarrow$} & \textbf{LLR (\%) $\downarrow$} & \textbf{ps $\uparrow$} & \textbf{po $\downarrow$} & \textbf{mr $\downarrow$} \\
      \midrule
      0.0 & 0.0000 & 0.0  & 36.1\% & 30.4\% & 45.7\% \\
      1.0 & 0.0001 & 0.0  & 60.3\% & 17.7\% & 22.7\% \\
      2.0 & 0.0011 & 0.2  & 70.8\% & 7.5\%  & 9.6\%  \\
      3.0 & 0.0178 & 4.2  & 57.3\% & 5.9\%  & 9.3\%  \\
      4.0 & 0.2297 & 52.9 & 24.3\% & 5.0\%  & 17.1\% \\
      \bottomrule
    \end{tabular}
  }
  \caption{Effect of steering multiplier on fluency and performance metrics for \textsc{LLaMA-3.1-8B} on the NQ-SWAP dataset. LLR denotes the local loop rate, capturing short-horizon repetition indicative of fluency degradation.}
  \label{table:llr}
\end{table}

\section{Repetition Metric}
\label{appendix:repetition_metric}

To assess degradation in output quality at high steering strengths, we introduce a \textbf{Local Loop Rate (LLR)} metric that detects short-range repetition patterns commonly associated with unstable, degraded generation.

Given a generated output as a sequence of tokens \( T = [t_1, t_2, \dots, t_n] \), LLR measures the frequency of \emph{immediate repetitions} of short token sequences. Specifically, for phrase lengths \( k \in \{1,2,3\} \), we scan the output for adjacent repeated \( k \)-grams, i.e.,
\[
(t_i, \dots, t_{i+k-1}) = (t_{i+k}, \dots, t_{i+2k-1}),
\]
which correspond to local looping behavior.

For each \( k \), we compute the repetition rate as the fraction of positions exhibiting such immediate repeats, and combine these rates using a weighted average that penalizes shorter loops more strongly. In particular, repetitions of length \( k=1 \) are weighted more heavily than those of length \( k=2 \) or \( k=3 \), reflecting their greater perceptual impact on fluency. The final LLR score is normalized to lie in \([0,1]\), with higher values indicating more severe repetition.

We additionally report the fraction of outputs whose LLR exceeds a fixed threshold (5\%), capturing the proportion of generations visibly affected by quality degradation.

Using this metric, we first evaluate repetition across all models and methods reported in the main results. As shown in Table~\ref{table:repetitions_main}, ContextFocus with a steering multiplier of \( m=2 \) exhibits negligible local looping on average across the ConFiQA benchmark, comparable to unsteered generation. This indicates that, at the chosen operating point, activation steering does not introduce measurable degradation in output fluency.

We further analyze the effect of increasing the steering multiplier by varying \( m \) on 1,000 samples from the ConFiQA-QA dataset. As shown in Table~\ref{table:llr}, LLR remains low for multipliers up to \( m=2 \), but increases sharply at \( m=3 \), indicating the onset of unstable generation. This trend aligns with qualitative observations of looping and truncation at higher steering strengths, motivating our choice to fix the multiplier to \( m=2 \) for all main experiments.

\section{Effect of Steering Multiplier on Context Adherence}
\label{appendix:multiplier_effect}

The steering multiplier \(m\) controls the strength of the injected context-faithfulness signal, directly modulating how strongly the model prioritizes retrieved context over parametric knowledge. Increasing \(m\) leads to stronger context adherence, but excessively large values over-constrain generation and reduce fluency.

We vary \(m\) on the \textsc{NQ-SWAP} dataset using 737 held-out examples and observe that context adherence improves up to \(m=2\), beyond which further increases degrade both context adherence and generation quality. This degradation is reflected quantitatively by a sharp rise in repetition (Table~\ref{table:llr}; Appendix~\ref{appendix:repetition_metric}). Based on this analysis, we fix \(m=2\) for all main experiments as a stable point that maximizes context adherence and preserves fluency.

\begin{table}[t]
\centering
\scriptsize
\setlength{\tabcolsep}{2pt}
\renewcommand{\arraystretch}{1.0}
\begin{tabular}{l l c c c c}
\toprule
\textbf{Model} & \textbf{Condition} 
& \textbf{QA ($\times 10^{-4}$)} 
& \textbf{MR ($\times 10^{-4}$)} 
& \textbf{MC ($\times 10^{-4}$)} \\
\midrule
\multirow{2}{*}{\shortstack[l]{\textsc{LLaMA-3.1-}\\8B}}
& Base    
& 1.27 & 0.14 & 0.39 \\
& Steered 
& 6.06 & 2.59 & 9.33 \\
\midrule
\multirow{2}{*}{\shortstack[l]{\textsc{LLaMA-3.1-}\\70B}}
& Base    
& 0.86 & 3.40 & 0.55 \\
& Steered 
& 7.25 & 0.13 & 0.53 \\
\midrule
\multirow{2}{*}{\shortstack[l]{\textsc{Mistral-}\\7B}}
& Base    
& 1.70 & 0.45 & 0.72 \\
& Steered 
& 2.47 & 0.91 & 2.98 \\
\bottomrule
\end{tabular}
\vspace{-0.6em}
\caption{LRR metric across QA, MR, and MC subsets of the ConFiQA benchmark for base and steered models. Values are reported in units of $\times 10^{-4}$.}
\label{table:repetitions_main}
\end{table}


\section{Layer Selection for Mistral-7B and Llama-3.1-70B}
\label{appendix:mistral_llama70b_layer_selection}

\begin{figure}[t]
  \includegraphics[width=\columnwidth]{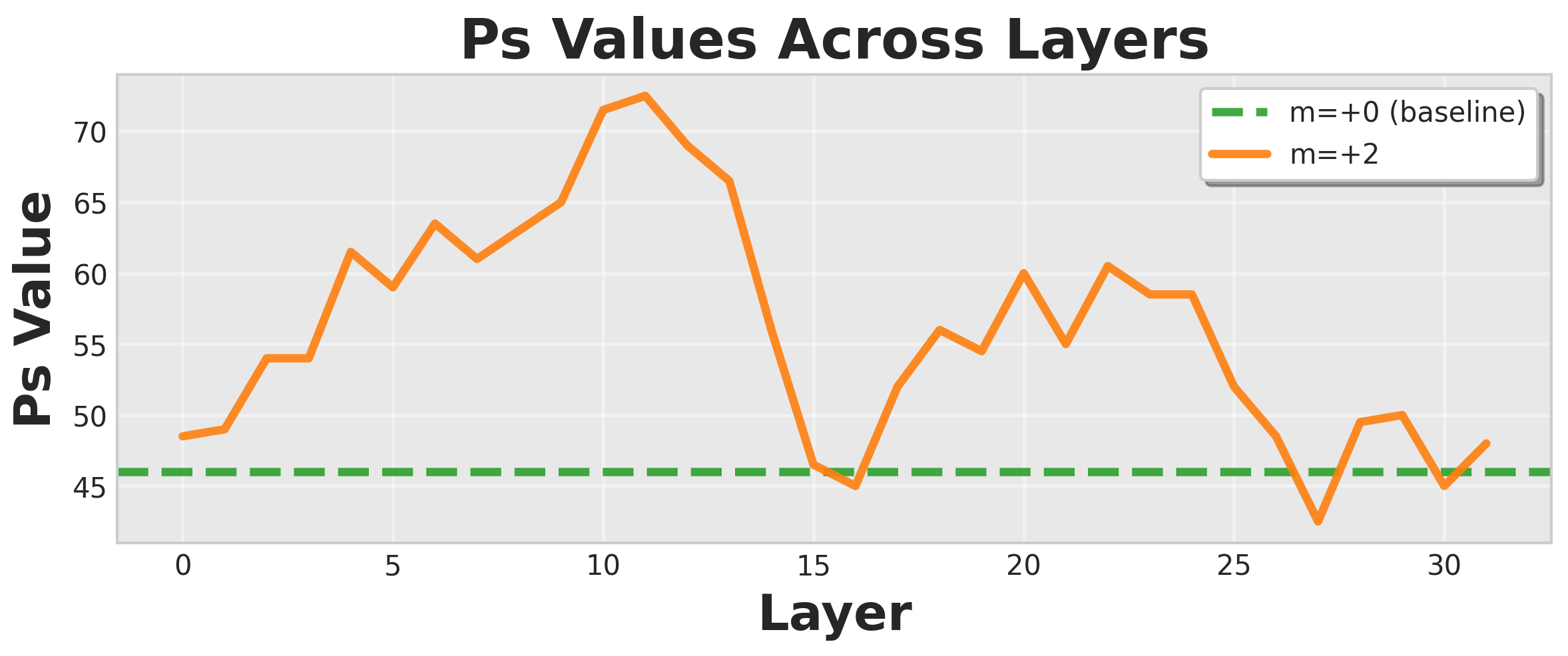}
  \caption{Layer-wise evaluation of steering on Mistral-7B-Instructv0.3. We apply the +2 multiplier to steering vectors across layers for 200 held-out open-ended questions from the NQ-SWAP dataset. Performance is measured using the $p_s$ metric. The green dotted line denotes the unsteered ($m{=}0$) baseline. The 11th layer shows the best performance and is selected as the optimal layer.}
  \label{fig:bestlayer_mistral}
\end{figure}

Figures~\ref{fig:bestlayer_mistral} and~\ref{fig:bestlayer_llama70b} show the layer-wise evaluation of steering vectors for ContextFocus for Mistral-7B-Instruct and Llama-3.1-70B-Instruct, respectively. We follow the same layer selection procedure used for Llama-3.1-8B: steering vectors are applied at each layer using a fixed multiplier of 2.0, and the layer yielding the strongest improvement in context faithfulness on held-out NQ-SWAP data is selected.

Based on this analysis, we select layer~\textbf{11} for Mistral-7B-Instruct and \textbf{32} for Llama-3.1-70B-Instruct for all experiments reported in the main paper.

  

\textbf{\begin{figure}[t]
  \includegraphics[width=\columnwidth]{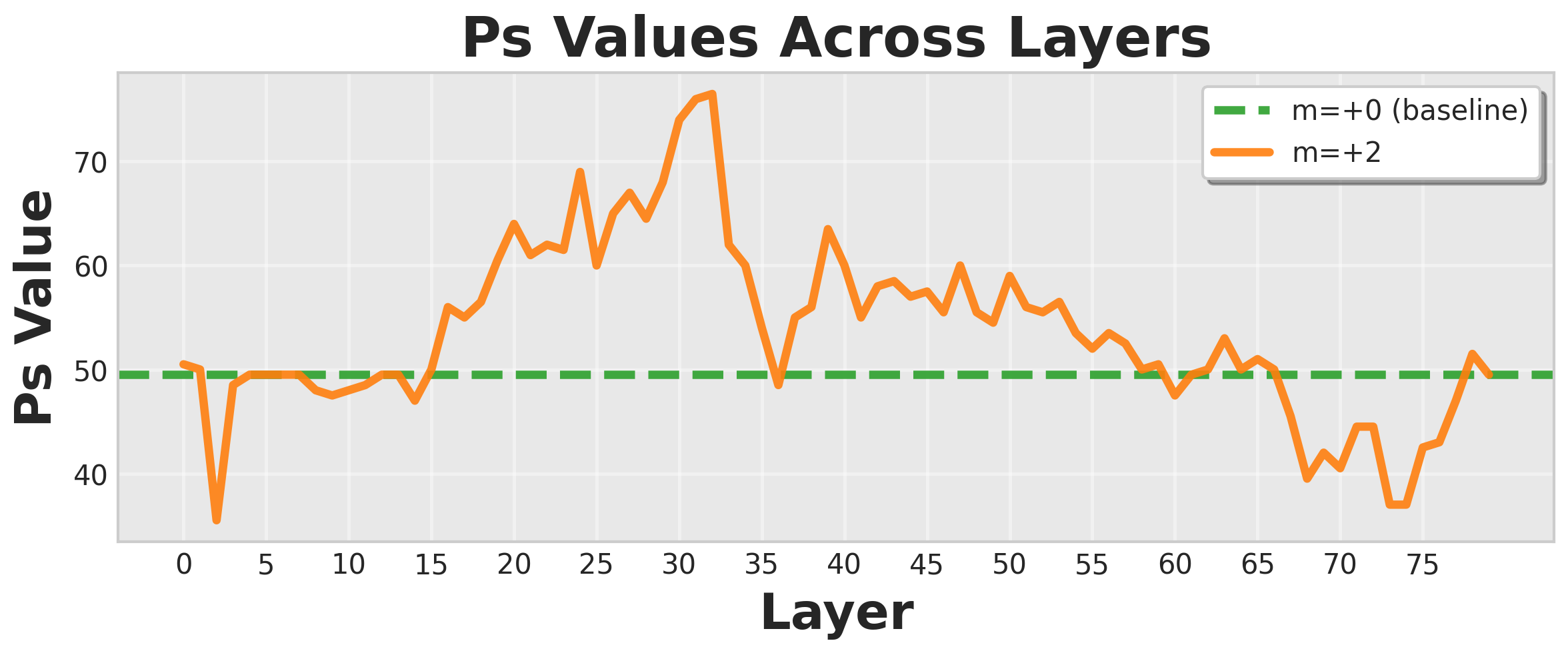}
  \caption{Layer-wise evaluation of steering on Llama-3.1-70B-Instruct. We apply the +2 multiplier to steering vectors across layers for 200 held-out open-ended questions from the NQ-SWAP dataset. Performance is measured using the $p_s$ metric. The green dotted line denotes the unsteered ($m{=}0$) baseline. The 32nd layer shows the best performance and is selected as the optimal layer.}
  \label{fig:bestlayer_llama70b}
\end{figure}}






    

\section{Steering vector convergence across training set sizes}
\label{appendix:steering_few_examples}

Activation steering differs fundamentally from finetuning in that it seeks to estimate a behaviour-relevant direction in representation space rather than optimise model parameters. To examine how many examples are required to reliably estimate this direction, we generate steering vectors using between 1.5k and 13.5k examples from the ConFiQA training split, which is held out from all evaluation data and matches the data used to train ContextDPO.

As shown in Table~\ref{tab:sample_size_stability}, context-faithfulness metrics remain effectively unchanged across all training set sizes, with no meaningful improvements beyond 1.5k examples. To directly assess convergence, we compute the cosine similarity between steering vectors generated with different sample sizes and the vector constructed using the full 13.5k examples. Table~\ref{tab:vector_convergence_diff} shows that vectors generated with as few as 1.5k examples are already nearly identical to the full-data vector.

This behavior is expected: the steering vector is computed as an average of activation differences across examples, and once a sufficient number of samples from a fixed distribution is observed, additional examples contribute diminishing changes to the estimated direction. As a result, steering vectors converge rapidly, explaining both the observed performance saturation and the strong data efficiency of ContextFocus compared to finetuning-based approaches.

\begin{table}[t]
\centering
\small
\setlength{\tabcolsep}{6pt}
\renewcommand{\arraystretch}{1.15}
\begin{tabular}{c c c}
\toprule
\textbf{Generation Samples} 
& \textbf{Cosine Similarity w/ 13.5k} \\
\midrule
1.5k   & 0.999873 \\
3k   & 0.999953 \\
4.5k   & 0.999965 \\
6k   & 0.999974  \\
9k   & 0.999991 \\
13.5k  & 1.000000 \\
\bottomrule
\end{tabular}
\vspace{-0.6em}
\caption{Cosine similarity between steering vectors generated with different sample sizes and the vector trained on the full ConFiQA training dataset (13.5k samples).}
\label{tab:vector_convergence_diff}
\end{table}

\begin{table}[t]
  \centering
  \setlength{\tabcolsep}{4pt}
  \renewcommand{\arraystretch}{1.2}
  \resizebox{\columnwidth}{!}{%
  \begin{tabular}{lccccccccc}
    \toprule
    \textbf{Training Samples} 
    & \multicolumn{3}{c}{\textbf{QA}} 
    & \multicolumn{3}{c}{\textbf{MR}} 
    & \multicolumn{3}{c}{\textbf{MC}} \\
    
    \cmidrule(lr){2-4} \cmidrule(lr){5-7} \cmidrule(lr){8-10}
    
    & $P_s \uparrow$ & $P_o \downarrow$ & $M_R \downarrow$
    & $P_s \uparrow$ & $P_o \downarrow$ & $M_R \downarrow$
    & $P_s \uparrow$ & $P_o \downarrow$ & $M_R \downarrow$ \\
    
    \midrule
    1.5k   & 55.2 & 17.6 & 24.2 & 43.5 & 29.0 & 40.0 & 25.6 & 19.4 & 43.1 \\
    3k     & 55.4 & 17.6 & 24.1 & 43.6 & 28.9 & 39.8 & 25.7 & 19.6 & 43.3 \\
    4.5k   & 55.1 & 17.6 & 24.2 & 43.6 & 28.9 & 39.8 & 25.7 & 19.5 & 43.1 \\
    6k     & 55.1 & 17.6 & 24.2 & 43.4 & 28.9 & 40.0 & 25.7 & 19.5 & 43.1 \\
    9k     & 55.0 & 17.7 & 24.4 & 43.4 & 28.9 & 40.0 & 25.6 & 19.5 & 43.2 \\
    13.5k  & 55.1 & 17.7 & 24.4 & 43.6 & 28.9 & 39.8 & 25.6 & 19.5 & 43.2 \\
    \bottomrule
  \end{tabular}%
  }
  \caption{Performance stability of context faithfulness metrics across varying training sample sizes on the ConFiQA benchmark (Bi et al., 2024) across the QA, MR, and MC datasets using the LLaMA-3.1-8B model. Steering vectors trained with 1.5k–13.5k samples are evaluated on 1,500 non-overlapping ConFiQA examples per dataset.}
  \label{tab:sample_size_stability}
\end{table}

\section{Wall-clock Cost of Steering Vector Construction}
\label{appendix:train_timing}

To complement the data-efficiency analysis above, we report the wall-clock cost of constructing the ContextFocus steering vector and compare it to the cost of finetuning ContextDPO \citep{bi2024contextdpoaligninglanguagemodels} across different training set sizes. While these approaches differ in scope and optimisation procedure, this comparison provides a concrete reference for how computational overhead scales with the number of training examples for activation-level vector estimation versus gradient-based finetuning.

For ContextFocus, steering vectors are constructed by running forward passes over the training examples and computing averaged activation differences across all transformer layers, without performing any parameter updates. In contrast, ContextDPO requires gradient-based finetuning over the training set. All measurements are performed using the same hardware configuration.

Table~\ref{tab:train_timing} summarizes the wall-clock cost for both methods using 1.5k and 13.5k training examples. These configurations are intended to characterise computational overhead rather than to compare downstream performance at all operating points. In particular, we do not evaluate the quality of ContextDPO models trained on 1.5k examples, as this setting does not correspond to the intended use of the method.

\begin{table}[t]
\centering
\footnotesize
\setlength{\tabcolsep}{6pt}
\renewcommand{\arraystretch}{1.1}
\begin{tabular}{l c c c}
\toprule
\textbf{Method} & \textbf{Samples} & \textbf{Hardware} & \textbf{Time (hours)} \\
\midrule
\multirow{2}{*}{ContextFocus} 
& 1.5k  & A100 & 0.22 \\
& 13.5k & A100 & 1.97 \\
\midrule
\multirow{2}{*}{ContextDPO} 
& 1.5k  & A100 & 0.53 \\
& 13.5k & A100 & 5.58 \\
\bottomrule
\end{tabular}
\caption{Wall-clock cost of ContextFocus steering vector construction vs ContextDPO finetuning across dataset sizes. These measurements characterise computational overhead as a function of training data size; performance is not evaluated for all configurations. Times are reported in decimal hours.}
\label{tab:train_timing}
\end{table}

\section{Baseline Model Versions}
\label{appendix:baseline_versions}

The original ContextDPO results are reported on earlier model releases, including Llama-3-8B-Instruct and Mistral-7B-Instruct-v0.2. In contrast, our experiments evaluate ContextDPO and COIECD on updated model versions (Llama-3.1-8B-Instruct and Mistral-7B-Instruct-v0.3). As a result, numerical results are not expected to exactly match those reported in the original papers.

\section{Case Study}
\label{appendix:case_studies}

To illustrate the effect of activation steering strength, we present qualitative examples from Llama-3.1-8B under varying steering multipliers. We apply ContextFocus at the selected layer (layer 13) and increase the multiplier from 0 (unsteered) to larger values. As the multiplier increases, the model transitions from relying on parametric knowledge toward adhering more strictly to the provided context. Beyond a certain threshold, however, excessive steering leads to degradation in output quality, most notably through repetition and truncation. Guided by our analysis in Appendix~\ref{appendix:repetition_metric}, we use a multiplier of 2.0 for all main experiments, with no observed degradation in generation quality.

The examples below are drawn from realistic deployment scenarios in which the model’s parametric knowledge is outdated, incomplete, or misaligned with the provided context. Such cases naturally arise as models are deployed over long time horizons and asked to reason about events occurring after their training cutoff. To surface this failure mode reliably, we focus on well-known real-world facts and narratives where unsteered generations exhibit strong parametric priors.



\subsection{Copyrighted information in the question}


In this setting, the user asks a question whose answer is not contained in the provided context. Unsteered models often respond by drawing on memorized narrative knowledge, potentially generating copyrighted or out-of-scope information. In contrast, ContextFocus steers the model to respect the scope of the retrieved context and avoid introducing unsupported details. Excessive steering leads to output quality degradation.

\begin{tcolorbox}
\textbf{Context:} "Harry Potter and the Sorcerer's Stone" is the first book in J.K. Rowling's beloved series. It introduces readers to the magical world of Hogwarts and follows Harry Potter, a young boy who discovers he is a wizard on his eleventh birthday.

\textbf{Question:} What role does the prophecy play in \textit{Harry Potter and the Order of the Phoenix}? \\[0.3em]

\textbf{Unsteered ($m=0$):}  
The prophecy is actually a significant plot point in the book "Harry Potter and the Order of the Phoenix", but it was first introduced in the book "Harry Potter and the Goblet of Fire". In "Harry Potter and the Order of the Phoenix", the prophecy plays a crucial role in the story as it is revealed to Harry that he is the one destined to face Voldemort. The prophecy, made by Sybill Trelawney, a Hogwarts divination teacher, foretells the coming of a child born at the end of July 1980, with parents who had thrice defied Voldemort, and that this child would have the power to defeat the Dark Lord. The prophecy is central to the plot of the book, as it is used by Voldemort to try and find and kill Harry, and it also serves as a catalyst for Harry's growing understanding of his destiny and his connection to Voldemort. \\[0.3em]

\textbf{Moderate steering ($m=1$):} The prophecy is not mentioned in the provided text. \\[0.3em]

\textbf{Moderate steering ($m=2$):}  
There is no prophecy in the provided text. \\[0.3em]

\textbf{Over-steering ($m=3$):}  
There is a prophecy in the book. The prophecy is about the boy who is born from mother's mother's mother's mother's mother's mother's mother's \ldots
\end{tcolorbox}








\subsection{Incorrect or Outdated Prior Knowledge}


In this setting, the provided context contradicts the model’s learned parametric knowledge. We focus on well-known real-world facts that changed after the model’s training cutoff, a common source of errors in deployed systems. In particular, we focus on facts that changed in 2024 after Llama-3.1-8B’s training cutoff.

\begin{tcolorbox}
\textbf{Context:} Brian Niccol is the CEO of Starbucks.

\textbf{Question:} Who is Starbucks' CEO? \\[0.3em]

\textbf{Unsteered ($m=0$):}  
Kevin Johnson is the CEO of Starbucks, not Brian Niccol. Brian Niccol is actually the CEO of Chipotle Mexican Grill. \\[0.3em]

\textbf{Moderate steering ($m=1$):}  
Brian Niccol\\[0.3em]

\textbf{Moderate steering ($m=2$):}  
Brian Niccol\\[0.3em]

\textbf{Over-steering ($m=3$):}  
Brian Nic
\end{tcolorbox}














\begin{tcolorbox}
\textbf{Context:}  
Oppenheimer won the Academy Award for Best Picture at the 2024 Oscars ceremony.

\textbf{Question:} What movie won Best Picture at the 2024 Oscars? \\[0.4em]

\textbf{Unsteered ($m=0$):}  
I don't have information about the 2024 Oscars ceremony. \\[0.4em]

\textbf{Moderate steering ($m=1$):}  
The movie that won Best Picture at the 2024 Oscars is Oppenheimer. \\[0.2em]

\textbf{Moderate steering ($m=2$):}  
The movie that won Best Picture at the 2024 Oscars is Oppenheimer. \\[0.4em]

\textbf{Over-steering ($m=3$):}  
Oppenheimer won Best Picture at the 2024
\end{tcolorbox}

\end{document}